\pdfoutput=1

\documentclass[11pt]{article}

\usepackage[]{acl}

\usepackage{times}
\usepackage{latexsym}

\usepackage{graphicx}
\usepackage{booktabs}
\usepackage{array}
\usepackage{subcaption}
\usepackage{multirow}
\usepackage{amsmath}
\usepackage{amsthm}
\theoremstyle{definition}
\newtheorem{define}{Definition}

\usepackage[T1]{fontenc}

\usepackage[utf8]{inputenc}

\usepackage{microtype}

\title{Manual-Guided Dialogue for Flexible Conversational Agents}

\author{\textbf{Ryuichi Takanobu$^1$, Hao Zhou$^2$, Yankai Lin$^2$, Peng Li$^2$, Jie Zhou$^2$, Minlie Huang$^1$} \\
  \small $^1$The CoAI group, DCST, Institute for Artificial Intelligence, State Key Lab of Intelligent Technology and Systems, \\
  \small $^1$Beijing National Research Center for Information Science and Technology, Tsinghua University, Beijing 100084, China \\
  \small $^2$Pattern Recognition Center, WeChat AI, Tencent Inc., China \\ 
  {\small \tt gxly19@mails.tsinghua.edu.cn, tuxzhou@tencent.com, aihuang@tsinghua.edu.cn} \\
}

\begin{document}
\maketitle
\begin{abstract}
How to build and use dialogue data efficiently, and how to deploy models in different domains at scale can be two critical issues in building a task-oriented dialogue system. In this paper, we propose a novel \textit{manual-guided dialogue} scheme to alleviate these problems, where the agent learns the tasks from both dialogue and manuals. The manual is an unstructured textual document that guides the agent in interacting with users and the database during the conversation. Our proposed scheme reduces the dependence of dialogue models on fine-grained domain ontology, and makes them more flexible to adapt to various domains. We then contribute a fully-annotated multi-domain dataset \textit{MagDial} to support our scheme. It introduces three dialogue modeling subtasks: \textit{instruction matching}, \textit{argument filling}, and \textit{response generation}. Modeling these subtasks is consistent with the human agent's behavior patterns. Experiments demonstrate that the manual-guided dialogue scheme improves data efficiency and domain scalability in building dialogue systems. \textbf{The dataset and benchmark will be publicly available for promoting future research}.
\end{abstract}

\section{Introduction}
\label{sec:intro}

Teaching machines to help humans accomplish their goals through conversation has long attracted many studies~\citep{gao2019neural,zhang2020recent}. 
Recent works~\citep{hosseini2020simple,peng2021soloist} heavily rely on fine-grained domain ontology~\citep{young2013pomdp} in building conversational agents to manage different scenarios, such as in-car navigation~\citep{eric2017key}, customer service~\citep{budzianowski2018multiwoz} and home automation~\citep{sciuto2018hey}. 

However, the construction of elaborate structured ontology, such as dialogue acts and slot-value pairs~\citep{budzianowski2018multiwoz,rastogi2020towards}, is extremely difficult and inefficient, as it requires both professional knowledge in dialogue systems and background knowledge across domains and services, which are hard to satisfy in practice. Furthermore, dialogue systems based on structured ontology knowledge are difficult to adapt to new domains and tasks, as the knowledge structure, \textit{e.g.} the definition of dialogue acts, may change significantly across domains, which is challenging to model for existing ontology-based dialogue systems.

\begin{figure}[!t]
    \centering
    \includegraphics[width=\linewidth]{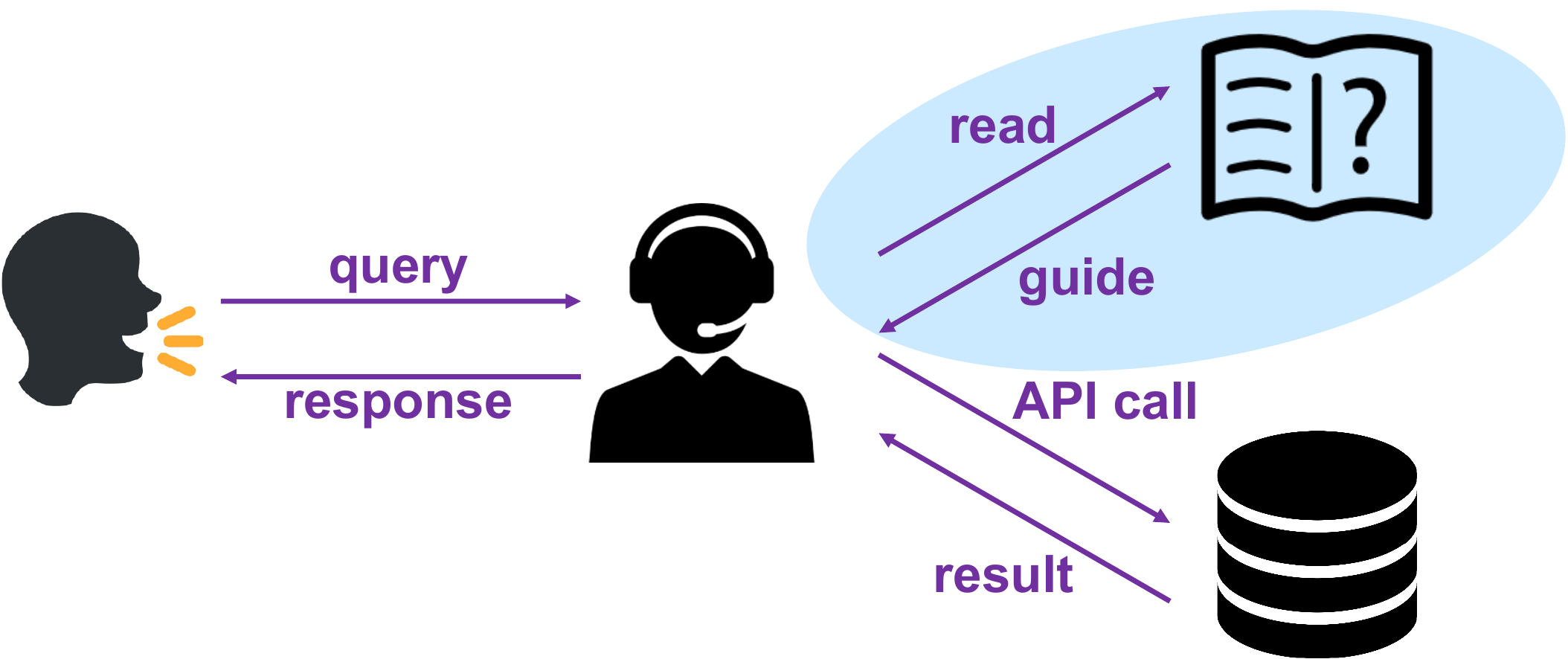}
    \caption{The overview scheme of a manual-guided dialogue system. The agent can access a textual manual (blue part) which provides the agent with guidance on completing user goals and making API calls during the conversation.}
    \label{fig:framework}
    \vspace{-0.3cm}
\end{figure}

In this paper, we propose the manual-guided dialogue scheme, where the agent can conduct task-oriented dialogues based on unstructured manuals instead of structured ontology. A manual is a collection of textual  \textit{instructions} that contains the task guidelines and database documentation an agent needs in the task-oriented dialogue scenario. In addition to making API calls to the database, the agent can read the manual to find the guide on how to act and respond whenever the agent converses with users (see Fig. \ref{fig:framework}). 
The manual-guided dialogue scheme allows developers to build unstructured manuals based on task-specific knowledge (see Sec.
\ref{sec:manual_collect}), which takes little effort compared to designing designing fine-grained structured ontology.
Once developers want to incorporate a new task, they only have to extend or modify the manuals, and then finetune the agent with few data to adapt to the new task.

To validate our hypothesis, we collect a six-domain task-oriented dialogue dataset \textit{MagDial} based on the proposed scheme, where 14 manuals are constructed with diverse language expressions to help and evaluate agents in understanding the unstructured task-specific knowledge. Three subtasks are introduced to facilitate training a manual-guided dialogue system following the above human behavior patterns: (1) \textit{instruction matching} subtask that finds the correct instructions, \textit{i.e.}, what the condition is and where the solution is, by scanning the manual to interpret user utterances, (2) \textit{argument filling} subtask that extracts the value spans from dialogue history according to the selected instructions to make accurate API calls, (3) \textit{response generation} subtask that generates appropriate responses conditioned on context and guided by the predicted outputs. Note that our data annotation process does not require well-defined ontology. 
A pipeline benchmarking three subtasks is also provided to demonstrate the usability of the manual-guided dialogue scheme. Results show that our pipeline improves data efficiency and domain scalability: Only hundreds of dialogue data are sufficient to yield effective dialogue models. Moreover, the trained models using our scheme perform well with unseen manuals.

Our contributions can be summarized as follows: 
\begin{itemize}
    \item We propose the scheme of manual-guided dialogue where the agent is guided by a manual consisting of task guidelines and database documentation during a task-oriented dialogue. 
    \item We collect a dialogue dataset MagDial following the proposed scheme and introduce three corresponding dialogue modeling subtasks: instruction matching, argument filling, and response generation. 
    \item We develop a flexible pipeline supported by the collected dataset. Extensive experiments demonstrate that the manual-guided dialogue scheme improves data efficiency and domain scalability.
\end{itemize}

\begin{figure*}[!t]
    \centering
    \includegraphics[width=0.9\linewidth]{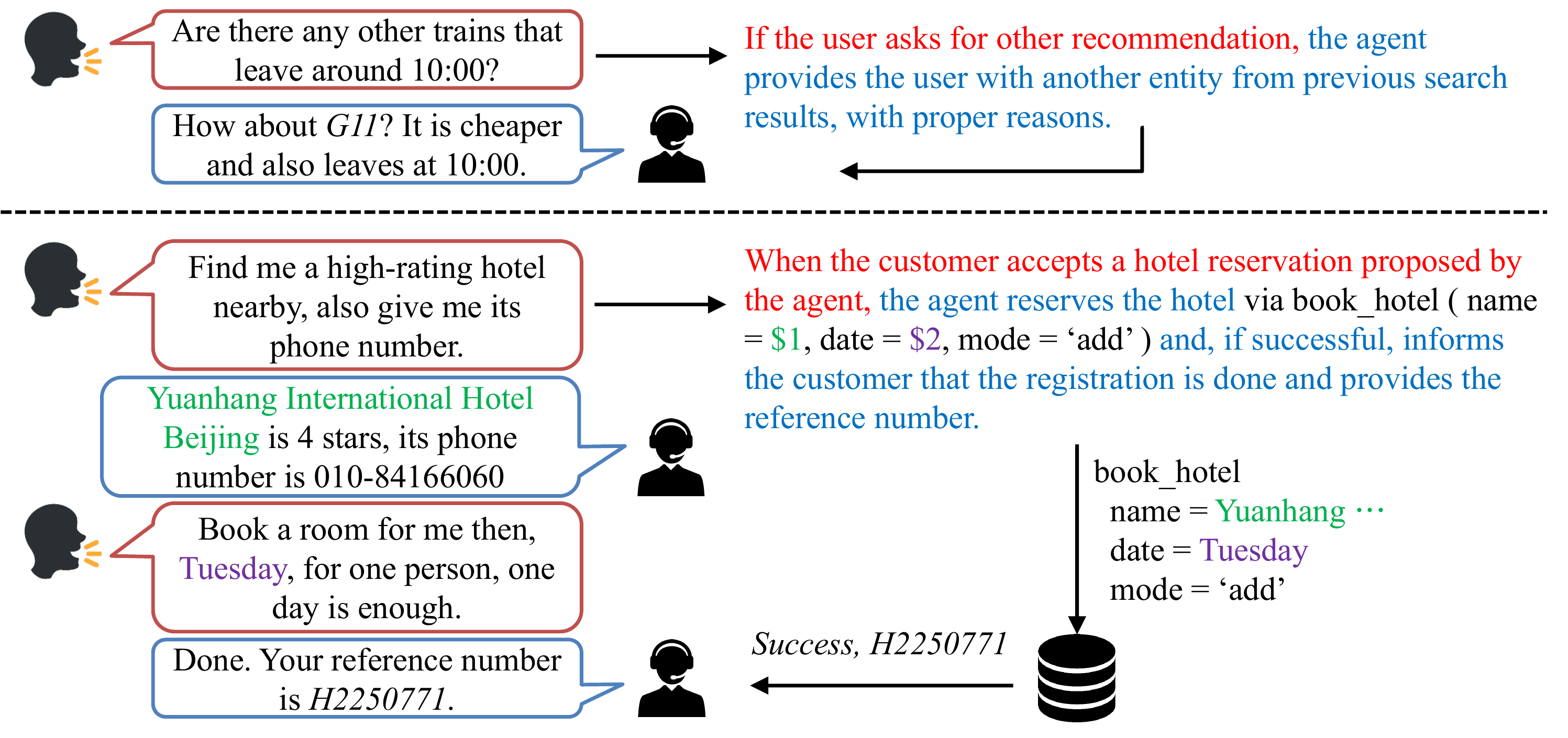}
    \caption{Several sample segments of conversation (left) in MagDial with selected instructions (right) in a manual. The paragraph marked in red/blue background color indicates the condition/solution part of an instruction. Each instruction may trigger an API call, and every API has its own input and output. As for argument filling, the agent detects the argument (\textit{e.g.}, Tuesday) from the dialogue context and recognizes its attribute type by reading the attribute mentions (\textit{e.g.}, date) in the selected instruction. The text spans to be extracted as an argument are marked in the same color with the corresponding attribute of APIs.}
    \label{fig:sample}
    \vspace{-0.3cm}
\end{figure*}

\section{Related Work}
\label{sec:related}

Most current techniques on task-oriented dialogue systems are data-intensive and require huge amounts of dialogue sessions to learn agent behaviors. Many large-scale corpora~\citep{budzianowski2018multiwoz,byrne2019taskmaster,peskov2019multi,rastogi2020towards} have been proposed to support different approaches. Training a pipeline system built on these datasets requires plenty of specialized labels for each module, \textit{e.g.}, language understanding~\citep{wang2018bi,goo2018slot} and dialogue state tracking~\citep{wu2019transferable,heck2020trippy}. However, collecting high-quality labeled data is expensive and time-consuming. Though pretrained end-to-end models~\citep{hosseini2020simple,peng2021soloist} achieve state-of-the-art (SOTA) performance, they predict dialogue states explicitly to inject semantics and constraints, which requires a large number of domain-specific labels as well. 

In addition to the data-hungry models, careful design of domain ontology is essential to dialogue modeling. In MultiWOZ~\citep{budzianowski2018multiwoz}, state trackers have to use heuristics by looking into the dataset and ontology to determine which slots should be categorical or span-based~\citep{gao2019dialogue,zhou2019multi}. Such high annotation complexity also causes substantial noise~\citep{eric2020multiwoz,zang2020multiwoz,han2020multiwoz}. 
Defining custom ontology in each work is another obstacle to rapid deployment. ABCD~\citep{chen2021action} designs more fine-grained intents and actions within a single domain. Others even introduce totally distinct ontology format, \textit{e.g.}, semantic frames~\citep{el2017frames} and meaningful representation~\citep{novikova2017e2e}. Due to the wide variety of structured ontology, methods on different datasets vary significantly in model structures and optimization.

Our work is inspired by recent studies on conversational reading comprehension~\citep{choi2018quac,saeidi2018interpretation} which investigates understanding a text passage to answer a series of interleaved user queries that appear in a conversation. To the best of our knowledge, the most relevant works to ours are as follows: SGD~\citep{rastogi2020towards} introduces natural language descriptions of intents and slots as input to support few-shot generalization in task-oriented dialogue modeling. Dialogue scenes in doc2dial~\cite{feng2020doc2dial} are grounded on various documents and the agent identifies relevant grounding content for information-seeking tasks. Some seed dialogue flows are first collected and then using user simulators for data augmentation to build a goal-oriented dialog agent efficiently via Alexa conversations~\citep{acharya2021alexa}.

However, our work includes, but is not limited to, the following three critical features beyond existing datasets: (1) The manual guides the agent in replying to users via conversation, and answers or responses cannot be derived from the manual itself. (2) The manual also guides making API calls to the database, rather than just explaining APIs by descriptions or comments. (3) Task-specific knowledge is represented in unstructured manuals. There are no annotations requiring domain expertise to define label categories or structures. (4) It is unnecessary to build a rule-based dialogue simulator that lacks scalability, nor train a model-based one with considerable extra labeled dialogue.

\section{Manual-Guided Dialogue}
\label{sec:MagDial}

In this section, we define some key concepts and formulate the task of manual-guided dialogue.

\begin{define}[Instruction]
An instruction is a sentence describing the contextual \textit{condition} of the current dialogue, the \textit{API call} description (optionally) an agent may need to obtain attributes from the database, and the corresponding possible \textit{solution} an agent may take to accomplish user goals.
\end{define}

\begin{define}[Argument]
An argument of an API is the value that needs to be provided (possibly optional) for API calls to obtain the API results.
\end{define}

In a manual-guided dialogue, the user and agent converse with each other at each turn $t$, obtaining the user utterance $U_t$ and agent response $R_t$. During the conversation, the agent can make API calls to retrieve relevant information from a database $DB$ including all candidate entities and corresponding attributes. Any API call holds the format of:
\begin{equation}
    func~( \text{attr}_1=\text{arg}_1, \text{attr}_2=\text{arg}_2, \cdots ).
\end{equation} 

Meanwhile, the agent can read a manual as guidance. A manual $M = \{I_1, I_2, \dots \}$ is a textual document comprised of the group of all instructions $I$ that instructs the agent how to interact with users and the database. Then the agent's goal is to learn a function $f$ that selects the relevant grounding instructions $I_t$, extracts the required arguments $A_t$ to be passed in API calls, and then generates an appropriate response $R_t$ given the dialogue history $D_t=\{U_1, R_1, \dots, U_{t-1}, R_{t-1}, U_t\}$, the database $DB$ and the manual $M$ as input:
\begin{equation}
    f : (D_t, DB, M) \rightarrow (I_t, A_t, R_t).
\end{equation}
Let $I_t = \{I_{t,1}, I_{t,2}, \dots I_{t,l}\}$ where $l$ denotes the number of instructions as multiple instructions can be selected at one dialogue turn. Similarly, $A_t = \{A_{t,1}, A_{t,2}, \dots A_{t,n}\}$ where $n$ denotes the number of arguments. Each argument is a text span extracted from $D_t$, since any categorical attribute where legal arguments are enumerable can be transformed into a span-based counterpart\footnote{An extracted argument of a categorical attribute that is outside the list of legal arguments can be mapped to the most relevant argument through similarity calculation or other methods, which is beyond the scope of this paper.}. Some instruction and argument examples are shown on the right side of Fig.~\ref{fig:sample}.

\section{MagDial Dataset}

This section introduces the MagDial dataset where the agent follows the manual-guided dialogue scheme during the conversation with users. Unlike traditional task-oriented dialogue corpus, the manuals are incorporated in this dataset to guide the agent to interact with users and the database, aiming to reduce the dependency on fine-grained structured ontology. The dataset, including manuals and dialogues, is collected in Chinese. 
Fig.~\ref{fig:sample} presents two dialogue segments from the dataset. From sample segments, we can observe that the selected instructions summarize the current dialogue situation (the condition part) and provide the agent with some guidance on the next action (the solution part). The agent may choose different $I_t$ from $M$ at a turn $t$ because multiple responses can be appropriate for the same $D_t$, leading to different valid dialogue policies towards task completion. 

Most parts of the manual include API calls (\textit{e.g.}, the second segment). The agent should follow the manual to extract arguments from the dialogue context to interact with the database properly. At each turn $t$, the agent selects relevant $I_t$ by scanning each instruction on $M$, and extracts $A_t$ to be passed in API calls by detecting text spans from $D_t$. When predicting text spans, the agent reads the $I_t$ to focus on the target arguments, and does not need to recognize its attribute type beforehand. This indicates that all the annotations used in our scheme are domain-agnostic\footnote{More technical details can be found in Sec.~\ref{sec:instr_match} and~\ref{sec:arg_fill}.}, thus improving the scalability across domains and tasks.

In the following text, we describe how we collect MagDial in three steps: (1) set up the background knowledge; (2) construct the manuals; (3) collect the dialogue and annotations.

\subsection{Data Setup}

We first set up the background knowledge needed in the task-oriented dialogue scenario, including the database, APIs, and user goals. Only the agent can interact with the database during the conversation, while the user has a user goal in mind and talks to the agent to accomplish the goal.

\paragraph{Database} 
For constructing the database $DB$, we crawled six-domain entities and corresponding attributes in Beijing from the Web. We choose the domains following MultiWOZ~\cite{budzianowski2018multiwoz} to give an intuitive comparison with existing datasets in experiments.
The statistics and full list of attributes are provided in Table \ref{tab:manual_db} and \ref{tab:attr_type} in Appendix due to the length limit.

\paragraph{API} 
There are four main API calls in a general database: \textit{find}, \textit{add}, \textit{edit} and \textit{delete} operations. In MagDial, we simplify the available user intents in dialogue to two categories: \textit{search} and \textit{booking}. A detected \textit{search} intent triggers a \textit{find} call that searches for relevant information, while a recognized \textit{booking} intent may trigger a(n) \textit{add} / \textit{edit} / \textit{delete} call that makes / changes / cancels a reservation based on the dialogue context. Note that API calls in MagDial do not edit the entities in the database. 
\textit{Carryover operation}\footnote{For example, if a user wants to find a Japanese restaurant with an API call $restaurant\_search ( \text{food} = \text{Japanese} )$ at first, and then he/she says that the restaurant should also be in the center of the city, the agent only has to make an API call $restaurant\_search ( area = center )$ and do not need to take ``Japanese'' as the argument again.} is always conducted in all API calls to carryover the arguments from previous turns, unless updated at the current turn.

\paragraph{User Goal}
A user goal $G=(C,R)$ is composed of constraints $C$ that indicate the specific conditions of the goal (\textit{e.g.}, find a \textit{Japanese} restaurant) and requests $R$ that represents what information or service the user seeks (\textit{e.g.}, require for the address of a restaurant). To construct user goals, we first sample the domains, with a max number of $4$ since too many domains in a goal make the dialogue unreal. Then we randomly sample the attributes as $C$ or $R$ accordingly. Note that one attribute should not be included in $C$ and $R$ simultaneously. For constraint attributes, we next sample the arguments from the database. Some rules are designed during the sampling process to eliminate unreasonable combinations.

\subsection{Manual Construction}\label{sec:manual_collect}

Although numerous manuals exist in real business scenarios, \textit{e.g.}, telephone customer service, we can hardly obtain these manuals and corresponding dialogues due to commercial restriction and data privacy. In order to provide open data resources for the research community and demonstrate the relatively simple process of manual construction, we decide to build the manuals $M$ from scratch.

\begin{figure}[!t]
    \centering
    \includegraphics[width=\linewidth]{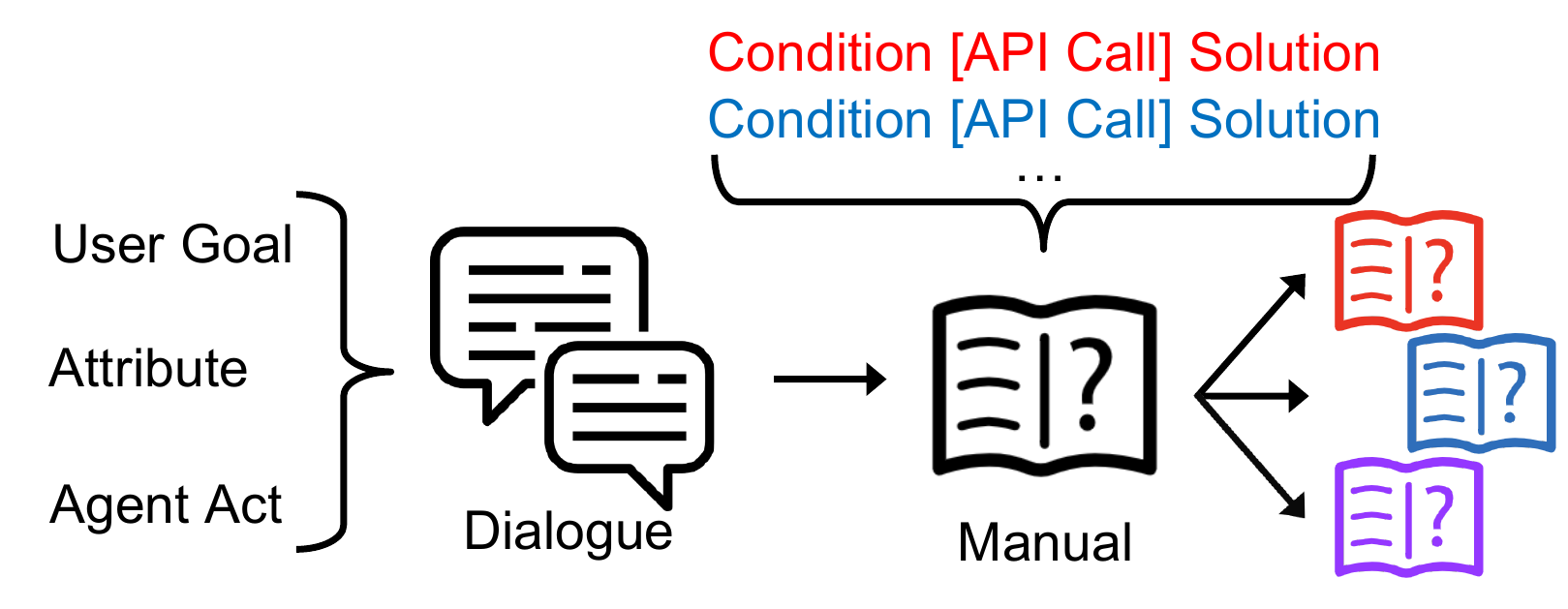}
    \caption{The process of manual construction.}
    \label{fig:manual_collect}
    \vspace{-0.3cm}
\end{figure}

An overview of the manual collection process is illustrated in Fig.~\ref{fig:manual_collect}. 
We first extract task-specific background knowledge, such as user goals and attributes, from the conversation in MultiWOZ~\cite{budzianowski2018multiwoz}. Then we manually design a small number of seed instructions, which contains a condition utterance describing user goals, an API call description (optionally) stating the method to obtain relevant attributes, and a solution utterance indicating the agent acts to complete user goals. Afterwards, we treat a manual $M$ as the composition of all the instructions $\{I\}$. The number of instructions in each domain is presented in Table \ref{tab:manual_db}.

To approximate the real business scenario that the style and expressions of manuals may be diverse across services, we further ask workers to paraphrase the seed instructions. In order to generate diverse paraphrases in terms of lexical and syntactic representation, each instruction is rewritten by one worker to avoid similar paraphrases, and self-BLEU~\cite{zhu2018texygen} score should be smaller than $0.8$. Finally, 14 paraphrases are created for each instruction, thus obtaining 14 manuals. 

\begin{table}[!ht]
    \centering
    \small
    \begin{tabular}{lrrr}
    \toprule
        Domain & Entity & Attribute & Instruction \\
    \midrule
        Attraction & 465 & 8 & 90  \\
        Hospital & 91 & 6 & 43 \\
        Hotel & 1,133 & 13 & 80 \\
        Restaurant & 951 & 12 & 114 \\
        Train & 1,022 & 12 & 133 \\
        Taxi\footnote[4] & - & 6 & 56 \\
    \bottomrule
    \end{tabular}
    \caption{Data statistics of the database and manuals.}
    \label{tab:manual_db}
\end{table}

\footnotetext[4]{We do not make up ``entities'' in \textit{Taxi} domain as anyone can book a taxi anytime and anywhere.}

\begin{table}[!ht]
    \centering
    \small
    \begin{tabular}{lrrr}
    \toprule
        Metric & Train & Dev & Test \\
    \midrule
        dialogues & 900 & 100 & 100 \\
        Turns & 5,350 & 562 & 652 \\
        Tokens & 294,956 & 30,633 & 37,571 \\
        Vocab & 5,593 & 2,187 & 2,298 \\
        Manuals & 10 & 10 & 4 \\
        Avg. dom. / dialogue & 2.42 & 2.36 & 2.73 \\
        Avg. turn / dialogue & 5.94 & 5.62 & 6.52 \\
        Avg. token / turn & 55.14 & 54.51 & 57.62 \\
        Avg. instr. / turn & 1.40 & 1.38 & 1.42\\
        Avg. arg. / turn & 1.17 & 1.22 & 1.19 \\
    \bottomrule
    \end{tabular}
    \caption{Statistics on dialogues. One turn here contains a user utterance and an agent response. The tokens are counted in Chinese.}
    \label{tab:dialogue}
    \vspace{-0.3cm}
\end{table}

\subsection{Dialogue Collection}

Following ~\citet{byrne2019taskmaster}, we adopt a self-play approach to collect the dialogue and annotations.
It has been demonstrated to render high-quality dialogue data efficiently and reasonably without too much unreality. The worker is asked to imagine the scenario where a user converses with an intelligent agent to complete the goal, and plays the user's role and the agent's alternately until the user goal is completed to collect the dialogue on his/her own. Unlike~\citet{byrne2019taskmaster} where workers write down the entire conversation using a single platform, the conversation environments for the ``user'' and ``agent'' are separate and different to ensure the workers switch the user/agent role during the conversation.

\paragraph{User Side}\label{sec:user}
At the beginning of the conversation, the worker is provided with a goal description and a table listing constraints and requests of the sampled goal. If a conversation ends without finishing the given goal, the goal will be put back to the task pool for other workers. To guarantee that the goal is properly expressed by the workers and successfully completed through conversation, the worker has to check the stated constraints at the current turn, or refill the value into the requests on the table at each turn. When all the constraints have been checked and all the requests have been filled, the task is completed.

\paragraph{Agent Side}\label{sec:agent}
When the worker acts as an agent, he/she tries to make API calls for users and respond based on the API results. For each conversation, the agent side is provided with a fixed manual. In order to get argument labels at the same time as dialogue collection, any API call is recorded. We use fuzzy string matching to annotate the span index of arguments in the dialogue context by looking into the recorded API calls during data post-processing. Moreover, the worker ought to select all the instructions used in the current turn as well. While it is prohibitive to select instructions from dozens of candidates at each turn, we provide the worker with a search interface to narrow down the instruction candidates during the annotation collection process. 
For all turns where workers select no instruction, we check to confirm that no instruction can be selected indeed, otherwise we regard the dialogue as a failure and ask the worker to repair it as a penalty.

\begin{figure*}[!t]
    \centering
    \begin{subfigure}[b]{0.28\textwidth}
        \centering
        \includegraphics[width=\linewidth]{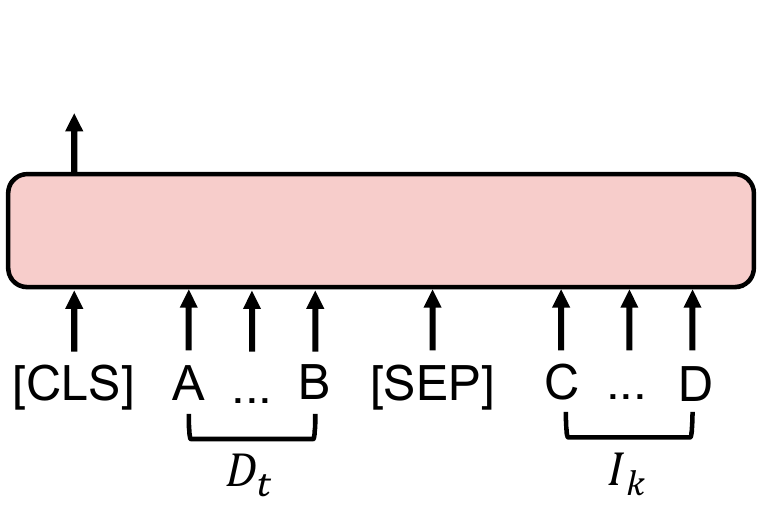}
        \caption{Instruction matching}
        \label{fig:instr_match}
    \end{subfigure}
    \begin{subfigure}[b]{0.315\textwidth}
        \centering
        \includegraphics[width=\linewidth]{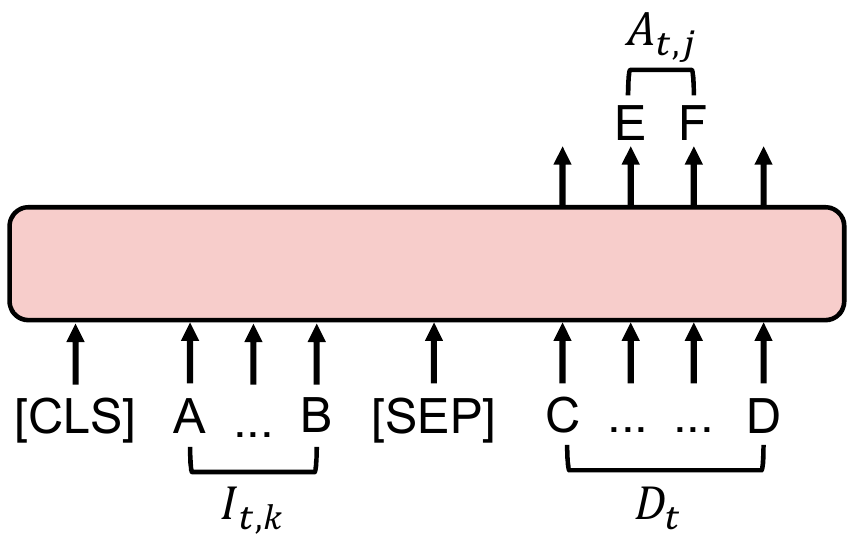}
        \caption{Argument filling}
        \label{fig:arg_fill}
    \end{subfigure}
    \begin{subfigure}[b]{0.385\textwidth}
        \centering
        \includegraphics[width=\linewidth]{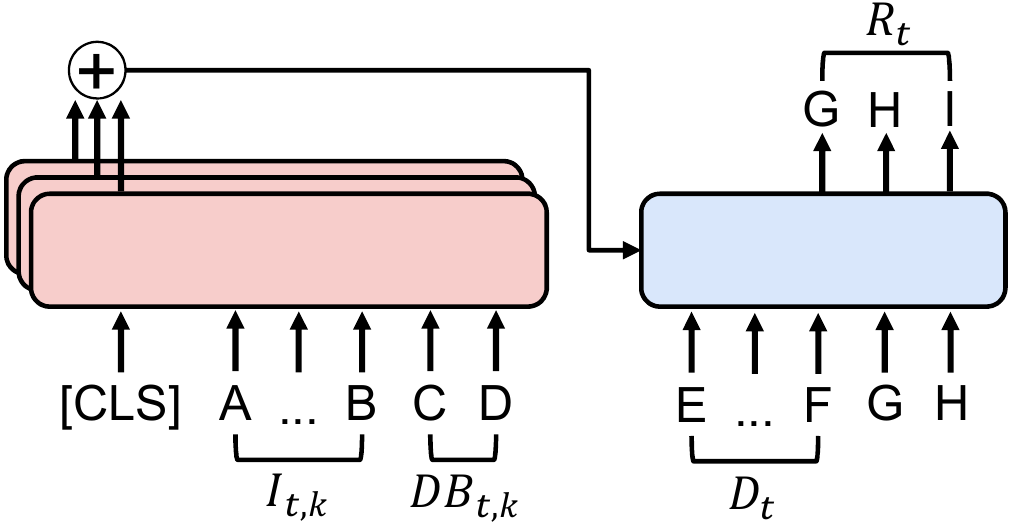}
        \caption{Response generation}
        \label{fig:res_gen}
    \end{subfigure}
    \caption{Benchmark modeling for each subtask. Red modules stand for encoders, and blue ones for decoders.}
    \label{fig:benchmark}
\end{figure*}
\subsection{Data Analysis}
\label{sec:data}

Some statistics of MagDial are presented in Table~\ref{tab:dialogue}. The dataset is built up from self-play conversations based on 1,100 distinct user goals. Each goal consists of 1 up to 4 domains, leading to 220/880 single-/multi-domain dialogues. The average number of turns is 5.97, while MultiWOZ~\cite{budzianowski2018multiwoz} consisting of 1 to 5 domains for each dialogue has an average 6.84 number of turns. This indicates the moderate complexity of our dataset. The maximum number of selected instructions at a turn is 10, and the maximum number of triggered arguments in an instruction is 2.

To measure the model's generalization, 10 out of 14 manuals are used for training/development set, and the rest 4 are for the test set, as shown in Table~\ref{tab:dialogue}. We collect 1,000 dialogues using 10 manuals, and the other 100 dialogues with the rest 4 manuals. The former 1,000 dialogues are randomly split into training and development sets. The 1,100 dialogues are constructed based on 1,100 distinct user goals, which leads to different semantics.

Fig. \ref{fig:manual_stat} in Appendix further illustrates some data distributions on the collected dialogues and manuals. The average number of sentence length is 22.27 and 32.41 for the user and agent side, respectively. After the manual paraphrasing process mentioned in Sec.~\ref{sec:manual_collect}, the average instruction length rises from 34.59 to 45.23, and its distribution becomes more normal. The number of instructions per turn shares a similar distribution with the number of arguments per turn since most instructions trigger one API call. Noticeably, neither an instruction is selected, nor an argument is triggered at a certain rate (around 20\%) of all dialogue turns since some turns only involve simple and generic dialogue behaviors such as greeting and farewells.

\section{Benchmarks}
\label{sec:benchmark}

To evaluate the benefits of a manual-guided dialogue scheme on improving data efficiency and domain scalability, we propose three subtasks with each a benchmark on the MagDial dataset to facilitate experimental analysis: \textit{instruction matching}, \textit{argument filling} and \textit{response generation}. Concerning benchmark models, we leverage the recent success of pretrained language models (PLMs)~\cite{devlin2019bert,liu2019roberta} on various NLP tasks and their potential towards general language understanding. An overview framework of the MagDial benchmarks is illustrated in Fig.~\ref{fig:benchmark}. All results are evaluated on the test set in subsequent subsections.

\begin{table}[!t]
    \centering
    \small
    \begin{tabular}{lcccc}
    \toprule
        Model & Acc. & Prec. & Rec. & F1 \\
    \midrule
        BERT-rec & 96.70 & 42.83 & 71.06 & 49.92 \\
        RoBERTa-rec & 96.73 & 43.07 & \textbf{75.33} & 50.74 \\
    \midrule
        BERT & 98.33 & 54.67 & 55.12 & 51.96 \\
        RoBERTa & \textbf{98.36} & \textbf{54.71} & 58.30 & \textbf{53.42} \\
    \bottomrule
    \end{tabular}
    \caption{Results of instruction matching.}
    \vspace{-0.3cm}
    \label{tab:instr_match}
\end{table}

\subsection{Instruction Matching}\label{sec:instr_match}

First of all, the agent selects applicable instructions $I_t$ from the manual $M$ given the dialogue history $D_t$. Since multiple instructions can be selected at a turn, we regard this as a multi-label sentence pair classification task. Each time the model encodes $D_t$ and a candidate instruction $I_k$, and predicts whether the condition part of $I_k$ matches the dialogue context. We use BERT-base~\cite{devlin2019bert} and RoBERTa-base~\cite{liu2019roberta} as the benchmark models. The model encodes the multi-turn $D_t$ as a long sequence, as shown in Fig.~\ref{fig:instr_match}. In terms of automatic metrics, we apply sentence-level accuracy to measure if each instruction $I_k$ in $M$ is correctly predicted, and turn-level F1 score to measure the alignment between the selected instructions $\hat{I}_t$ and ground truth instructions $I_t$.

The results of instruction matching are presented in Table~\ref{tab:instr_match}. We adopt two strategies during model optimization. One is to select the model with the highest recall on the development set during training iterations ($-rec$), the other uses F1 score as usual. 
It shows that high-recall targeted BERT and RoBERTa can achieve over 70\% recall with the unseen manuals. We can also observe that a standard PLM can achieve above 50\% F1 score and 98\% accuracy. 

\subsection{Argument Filling}\label{sec:arg_fill}

To enable the agent to interact with $DB$, we next need to extract relevant arguments $A_t$ from $D_t$ to be filled into the APIs triggered by the selected $I_t$. We regard this as a sequential tagging task with the \texttt{BIO} (Begin, Inside, Outside) format~\cite{ramshaw1999text}. Considering that there can be multiple attributes in one API call, we extend the \textit{Begin} and \textit{Inside} tags with an index indicating the attribute order in API to address the multi-argument extraction issue. Taking the second segment in Fig.~\ref{fig:sample} as an instance, the tags for \textit{Yuanhang International Hotel Beijing} and \textit{Tuesday} are \texttt{B-1, I-1, I-1, I-1} and \texttt{B-2} respectively, and \texttt{O} for other tokens. Here we apply the same benchmarks and similar encoding methods as the ones in the previous subtask, as shown in Fig.~\ref{fig:arg_fill}. As for automatic metrics, given a certain $I_{t,k}$, token-level accuracy on tag classification and turn-level F1 score between extracted arguments $\hat{A}_t$ and ground truth arguments $A_t$ are used.

\begin{table}[!t]
    \centering
    \small
    \begin{tabular}{lcccc}
    \toprule
        Model & Acc. & Prec. & Rec. & F1 \\
    \midrule
        BERT & 98.58 & 91.57 & 90.69 & 91.13 \\
        RoBERTa & \textbf{99.14} & \textbf{92.34} & \textbf{92.57} & \textbf{92.46} \\
        RoBERTa (w/o Manual) & 91.23 & 56.96 & 64.59 & 60.54 \\
    \bottomrule
    \end{tabular}
    \caption{Results of argument filling.}
    \label{tab:arg_fill}
    \vspace{-0.3cm}
\end{table}

Results on Table~\ref{tab:arg_fill} demonstrate that both BERT and RoBERTa can achieve a relatively high f1 score (over 98\%/90\% on the token/turn level) on argument filling thanks to our domain-agnostic annotations. Though there are many domains and attributes, the model's output space can be restricted to an extremely low number by the nature of the proposed tagging scheme (|\texttt{BIO}| $\times$ |\textit{index}|). The high performance also indicates that the trained models can correctly recognize different attributes on manuals and associate them to the proper arguments in the dialogue context. 
Noticeably, the F1, precision, and recall scores decrease significantly if the model cannot access manuals (the last line). 
It implies that manual-guided dialogue can teach the agent to interact with the database through reading manuals.
\subsection{Response Generation}

\begin{table}[!t]
    \centering
    \small
    \begin{tabular}{lccc}
    \toprule
        Model & PPL$\downarrow$ & BLEU$\uparrow$ & AER$\downarrow$ \\
    \midrule
        TT + BERT & \textbf{2.98} & 44.18 & 29.24 \\
        TT + RoBERTa & \textbf{2.98} & \textbf{45.02} & \textbf{28.12} \\
        TT (w/o Manual) & 3.45 & 42.15 & 32.22 \\
    \bottomrule
    \end{tabular}
    \caption{Results of response generation.}
    \label{tab:res_gen}
    \vspace{-0.3cm}
\end{table}

Finally, we aims to generate appropriate $R_t$ given all the $D_t$, $I_t$, and API results $DB_t$. We model this subtask as a conditional language generation task. TransferTransfo (TT)~\cite{wolf2019transfertransfo}, a generative pretrained dialogue system, is used as the benchmark generator. Similarly, BERT and RoBERTa are adopted as manual encoders. Since there can be multiple instructions in $I_t$ and therefore multiple results in $DB_t$, we apply a scaled dot-product attention mechanism Attention$(Q,K,V)$~\cite{vaswani2017attention} to get a hidden representation over $I_t$ and $DB_t$
, as shown in Fig.~\ref{fig:res_gen}:
\begin{align}
    \mathbf{H}_t &= \text{Encoder}([I_t;DB_t]), \notag \\
    \mathbf{z}_{t,j} &= \text{Generator}([D_t;R_{t,<j}]), \\
    \mathbf{z}_{t,j} &= \mathbf{z}_{t,j} + \text{Attention}(\mathbf{z}_{t,j}, \mathbf{H}_t, \mathbf{H}_t), \notag
\end{align}
where $\mathbf{z}_{t,j}$ is the hidden vector used to compute the $j$-th token's probability of the generated sample $R_t$. 
To evaluate the guidance of manuals on response generation, we also apply a single TT as a baseline for ablation study, which directly encodes the concatenated sequence of $DB_t$ and $D_t$ as conditions. For a fair comparison, the encoder's parameters are frozen during optimization for those methods encoding manuals. 
We adopt Perplexity (PPL) and BLEU score (geometric mean of 1$\sim$4-gram)~\cite{papineni2002bleu} to measure the generation quality. Furthermore, Argument Error Rate (AER) is introduced to evaluate whether the agent replies to the user with meaningful information by including return results $DB_t$ in the generated response $\hat{R}_t$.

Table~\ref{tab:res_gen} shows that the agent gives better responses after using manuals. A clear decrease on PPL and a distinct increase on BLEU are realized by encoding manuals as additional conditions. With the guidance of manuals, the agent achieves lower AER, therefore the generated response offers more accurate information. 

\section{Experiments}
\label{sec:experiment}

In this section, we further demonstrate the ability of a conversational agent grounded on manual-guided dialogue in data efficiency and manual/domain scalability.
We focus on \textit{joint instruction matching and argument filling subtasks} in this section, since these serve for the understanding of background knowledge in task-oriented dialogue and thus is essential for downstream tasks.

\subsection{Data Efficiency}\label{sec:data_size}

\begin{figure}[!t]
\centering
    \includegraphics[width=0.49\linewidth]{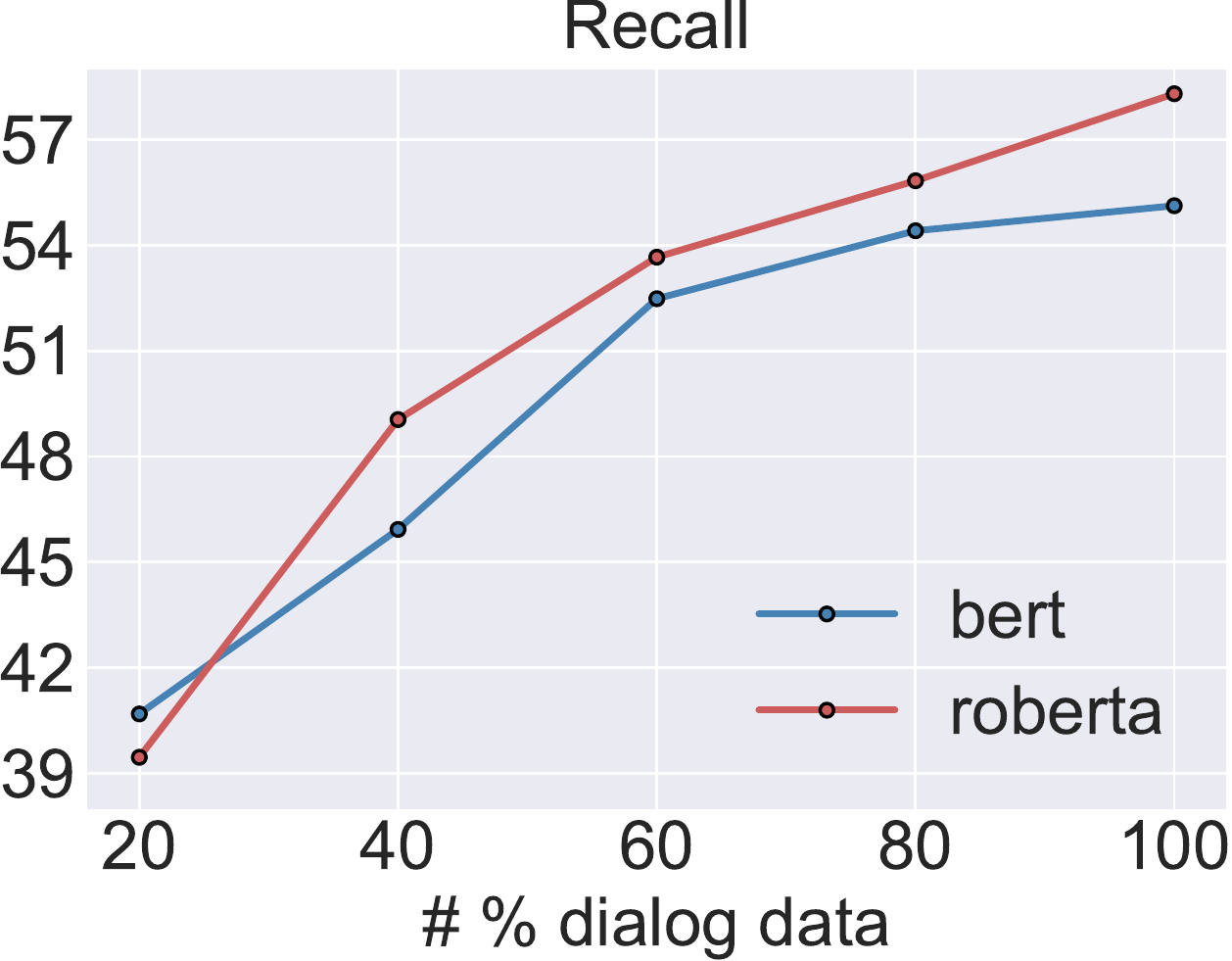}
    \includegraphics[width=0.49\linewidth]{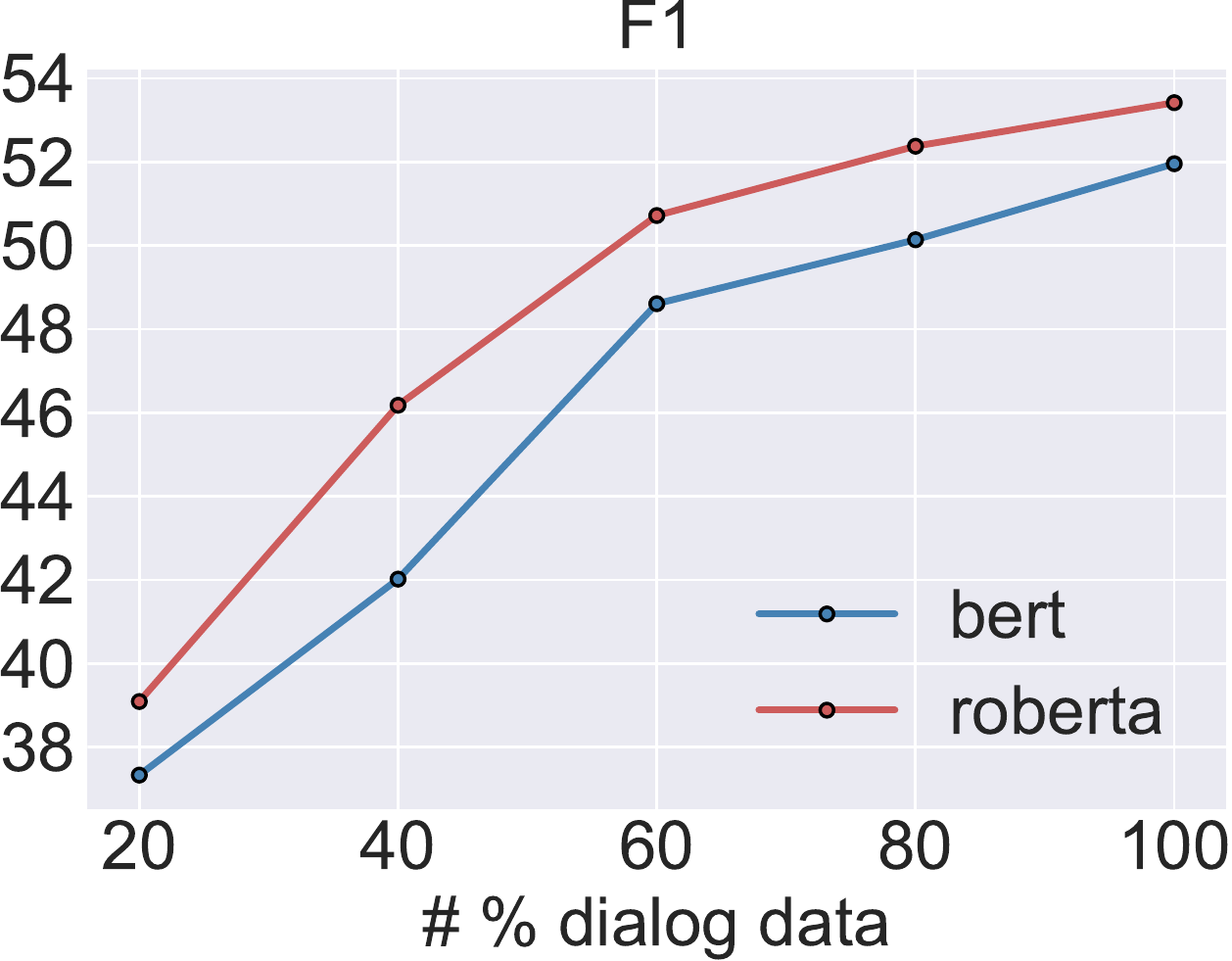}
\caption{Results w.r.t. the size of training data.}
\label{fig:data_size}
\vspace{-0.3cm}
\end{figure}

Fig.~\ref{fig:data_size} examines the agent's dependency on training data's size. We train the agent with 20\%, 40\%, 60\%, 80\% of the training set, respectively. We can observe that the agent performs better with the growth of training data size. Moreover, both recall and F1 scores increase rapidly before 60\% data.
We also found that models trained on about 500 dialogues (60\% of training data) can already reach a 0.5 F1 score, which is close to the performance of models trained on full data. These show that manual-guided dialogue scheme has a strong few-shot ability that can develop the task-oriented dialogue system with less training data.

\begin{figure}[!t]
\centering
    \includegraphics[width=0.49\linewidth]{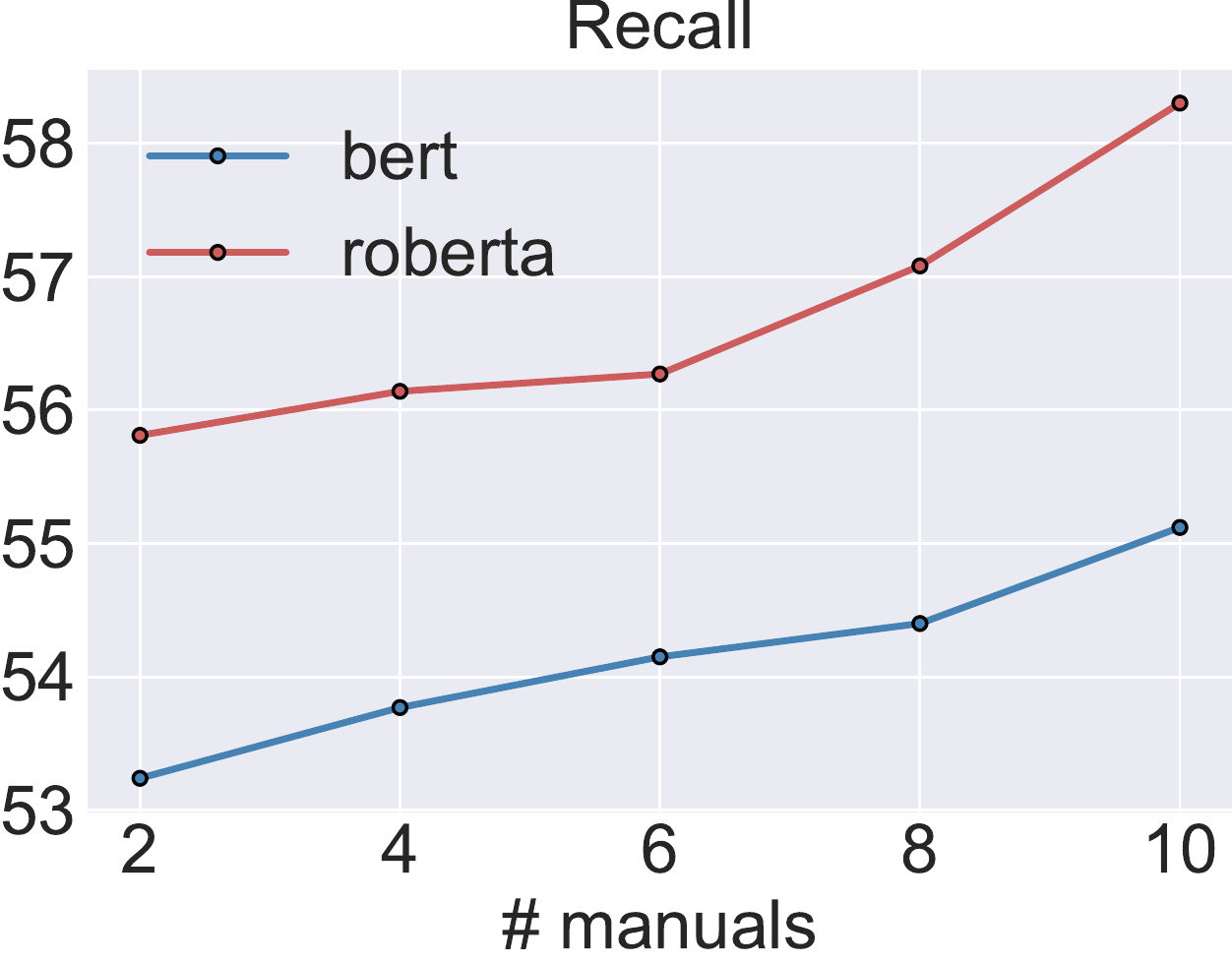}
    \includegraphics[width=0.49\linewidth]{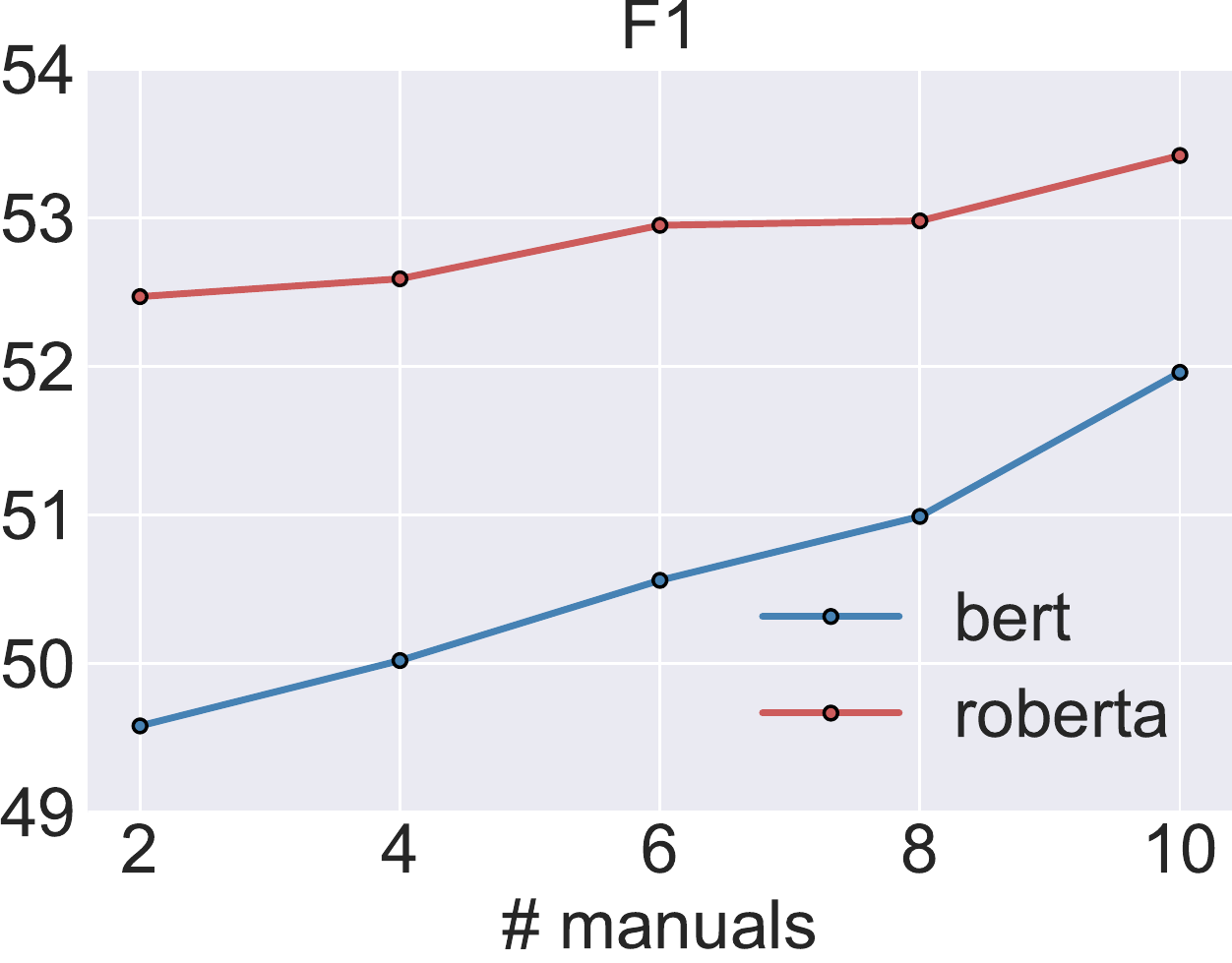}
\caption{Results w.r.t. the number of manuals in training data.}
\label{fig:manual_num}
\vspace{-0.3cm}
\end{figure}

\subsection{Manual Scalability}

Fig.~\ref{fig:manual_num} studies the agent's dependency on the number of training manuals to help it adapt to a different manual \textit{within domains}. We optimize the agent by using only part of (2, 4, 6, or 8) manuals in training and development sets. We find that using more manuals boosts the agent's performance as well, but surprisingly, conversational agents can quickly obtain around 0.5 F1 scores with only a couple of manuals. This indicates that the proposed scheme has relatively good scalability within domains. 

\subsection{Domain Scalability}

\begin{table}[!t]
    \centering
    \small
    \begin{tabular}{l@{~~~~}c@{~~~~}c@{~~~~}c@{~~~~}c@{~~~~}c}
    \toprule
        Domain & Att. & Hos. & Hot. & Res. & Tra. \\
    \midrule
        Full & \underline{48.49} & \textbf{54.44} & 54.04 & 49.90 & 50.40 \\
        \ \ - Attraction & \underline{40.63} & 45.85 & \textbf{48.80} & 45.77 & 42.61 \\
        \ \ - Hospital & \underline{43.18} & 48.42 & \textbf{49.37} & 44.79 & 45.37 \\
        \ \ - Hotel & 37.57 & \textbf{39.36} & \underline{37.23} & 36.55 & 37.59 \\
        \ \ - Restaurant & \underline{42.95} & 46.00 & \textbf{47.77} & 45.34 & 45.99 \\
        \ \ - Train & 43.12 & \textbf{45.79} & 45.18 & 45.41 & \underline{41.28}\\
    \bottomrule
    \end{tabular}
    \caption{F1 score of BERT by excluding one domain from training data (row) and tested on the dialogues that involve a certain domain (column). For each row, the best score is \textbf{bolded}, and the worst score is \underline{underlined}.}
    \label{tab:task_domain}
    \end{table}

To further explore the agent's scalability \textit{across domains}, we exclude one domain from training and development sets by removing the dialogues which involve that domain. As expected, Table~\ref{tab:task_domain} reveals a slight decline in all domains, whichever domain is excluded. Interestingly, the model's performance on the unseen domain varies from domain to domain (good for \textit{Hospital} and \textit{Restaurant}, poor for \textit{Attraction} and \textit{Train}). This implies manual-guided dialogue has moderate domain scalability across domains.

\section{Conclusion}
\label{sec:conclusion}

We propose a manual-guided dialogue scheme where the agent learns the tasks from both dialogues and manuals. The manual contains the task guideline to help the agent understand the context and converse with users, and also contains the database documentation to assist the agent in interacting with the database. Through reading the unstructured textual manuals, conversational agents can accomplish task goals more efficiently and are more flexible to handle multiple domains and make API calls without relying on the elaborate structured ontology. Based on the proposed scheme, we collect the MagDial dataset and introduce three subtasks to build a benchmark pipeline grounded on it. Extensive experiments demonstrate that the manual-guided dialogue scheme improves data efficiency and domain scalability.

\bibliography{reference}
\bibliographystyle{acl_natbib}

\appendix

\section{Appendices}

\subsection{Data Statistics}

\begin{table*}[!t]
    \centering
    \small
    \begin{tabular}{cm{13.5cm}}
    \toprule
        Attribute & address, area, arrive, car, choice, class, day, department, departure, destination, facility, food, id, leave, name, people, phone, postcode, price, reference num., score, station, star, stay, time, type \\
                \bottomrule
    \end{tabular}
    \caption{Full list of attributes extracted in the database, and used in manuals.}
    \label{tab:attr_type}
\end{table*}

\begin{figure*}[!t]
    \centering
    \includegraphics[width=0.325\textwidth]{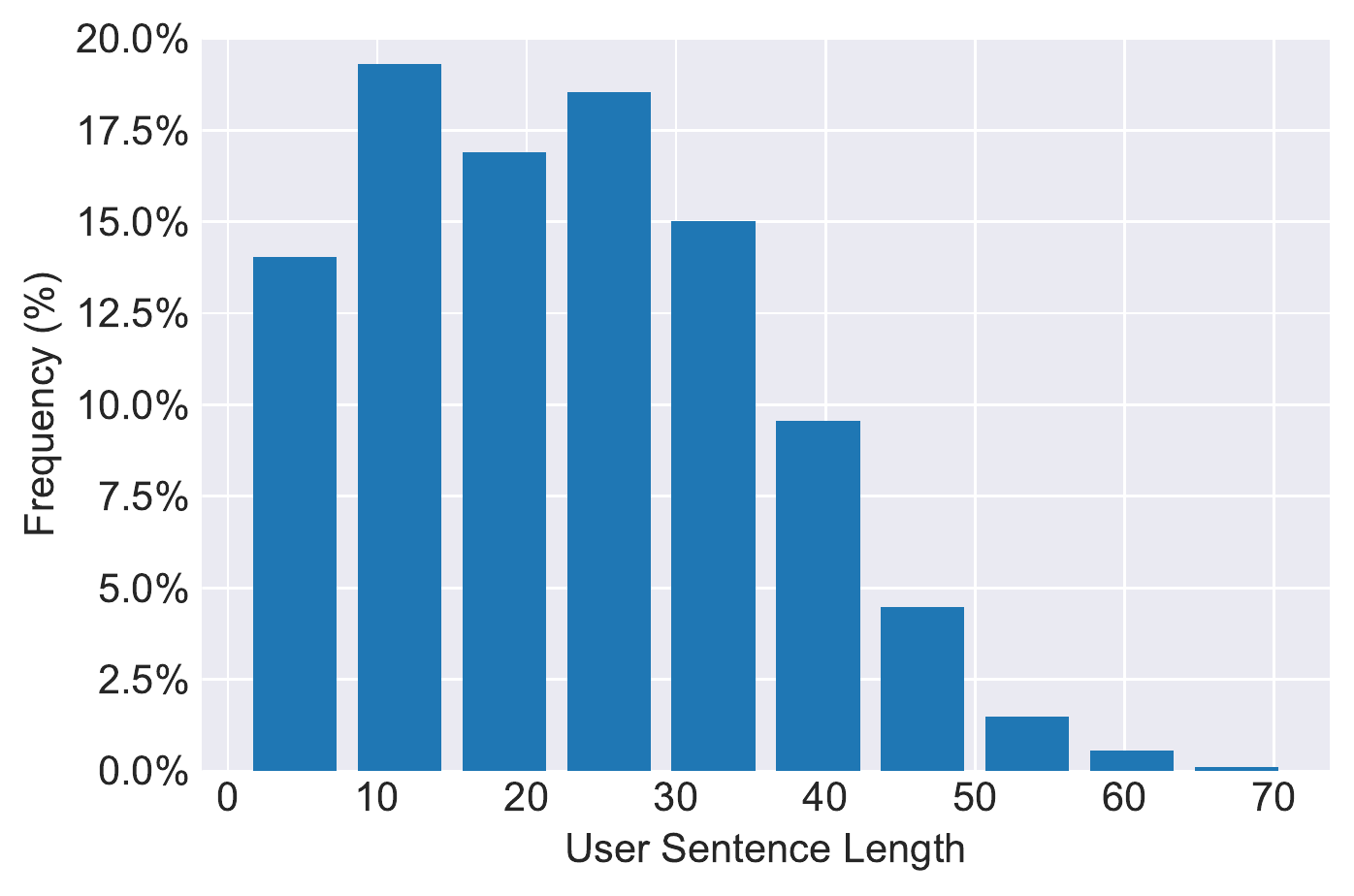}
    \includegraphics[width=0.325\textwidth]{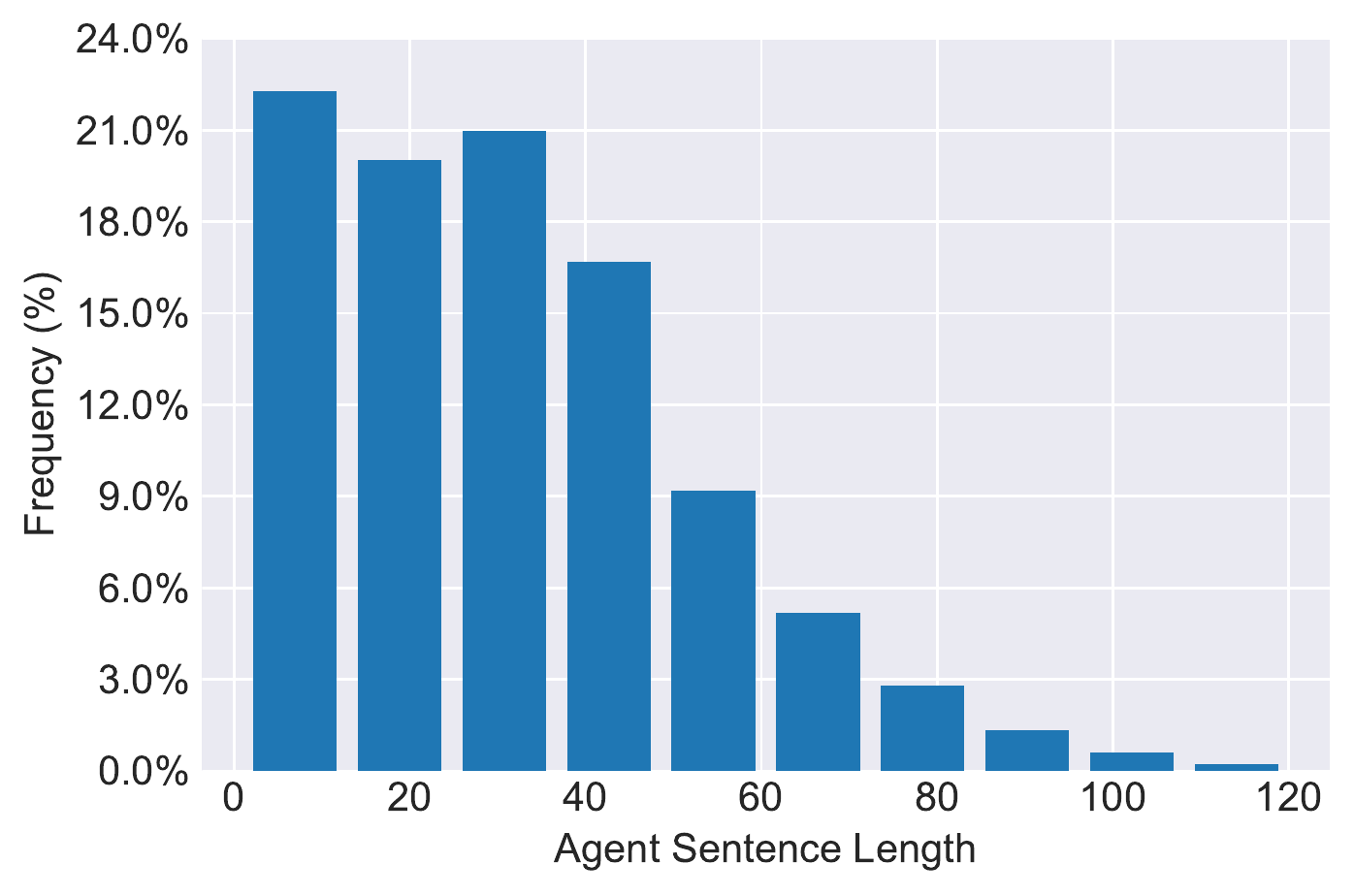}
    \includegraphics[width=0.325\textwidth]{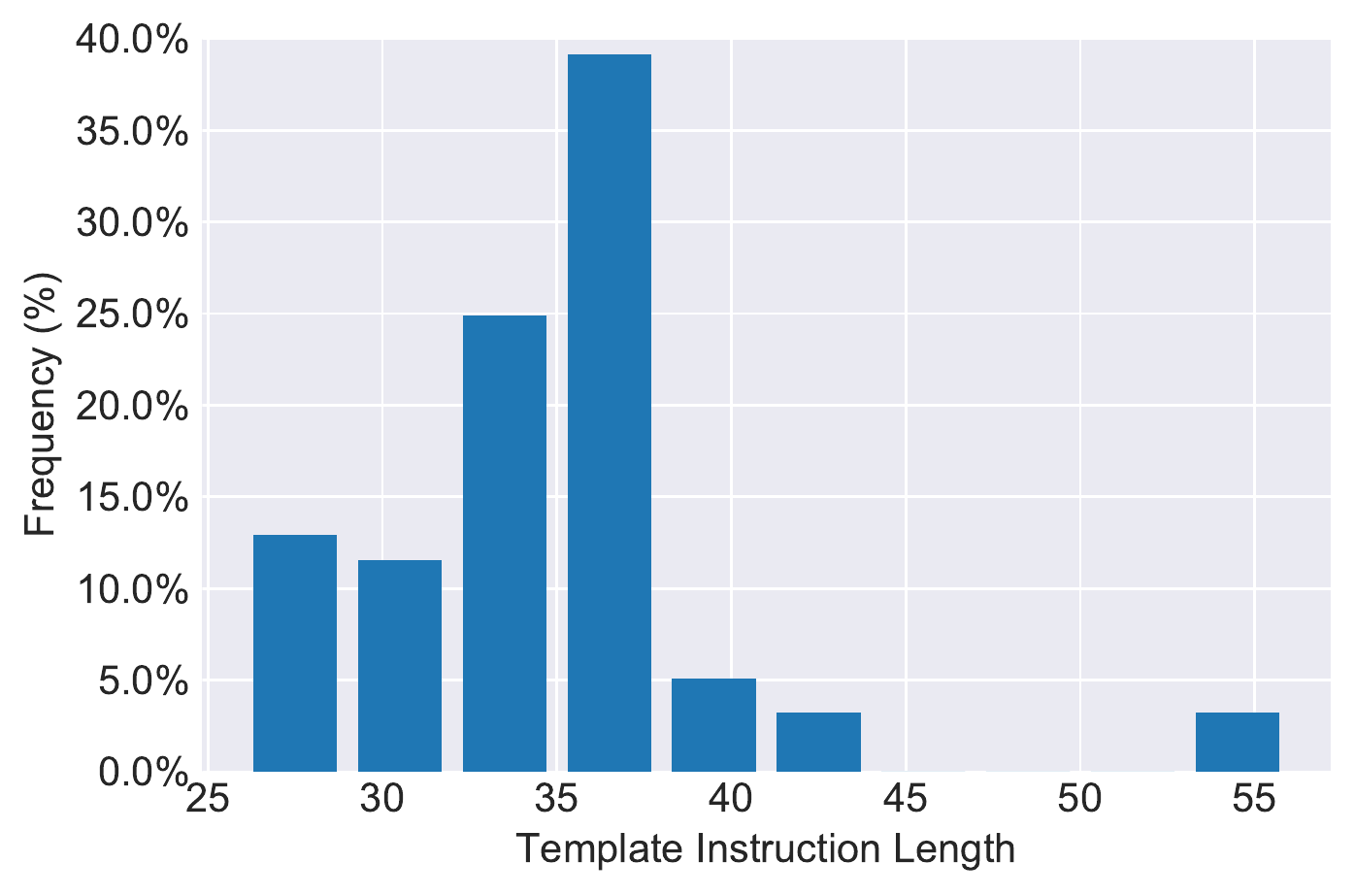}
    \includegraphics[width=0.325\textwidth]{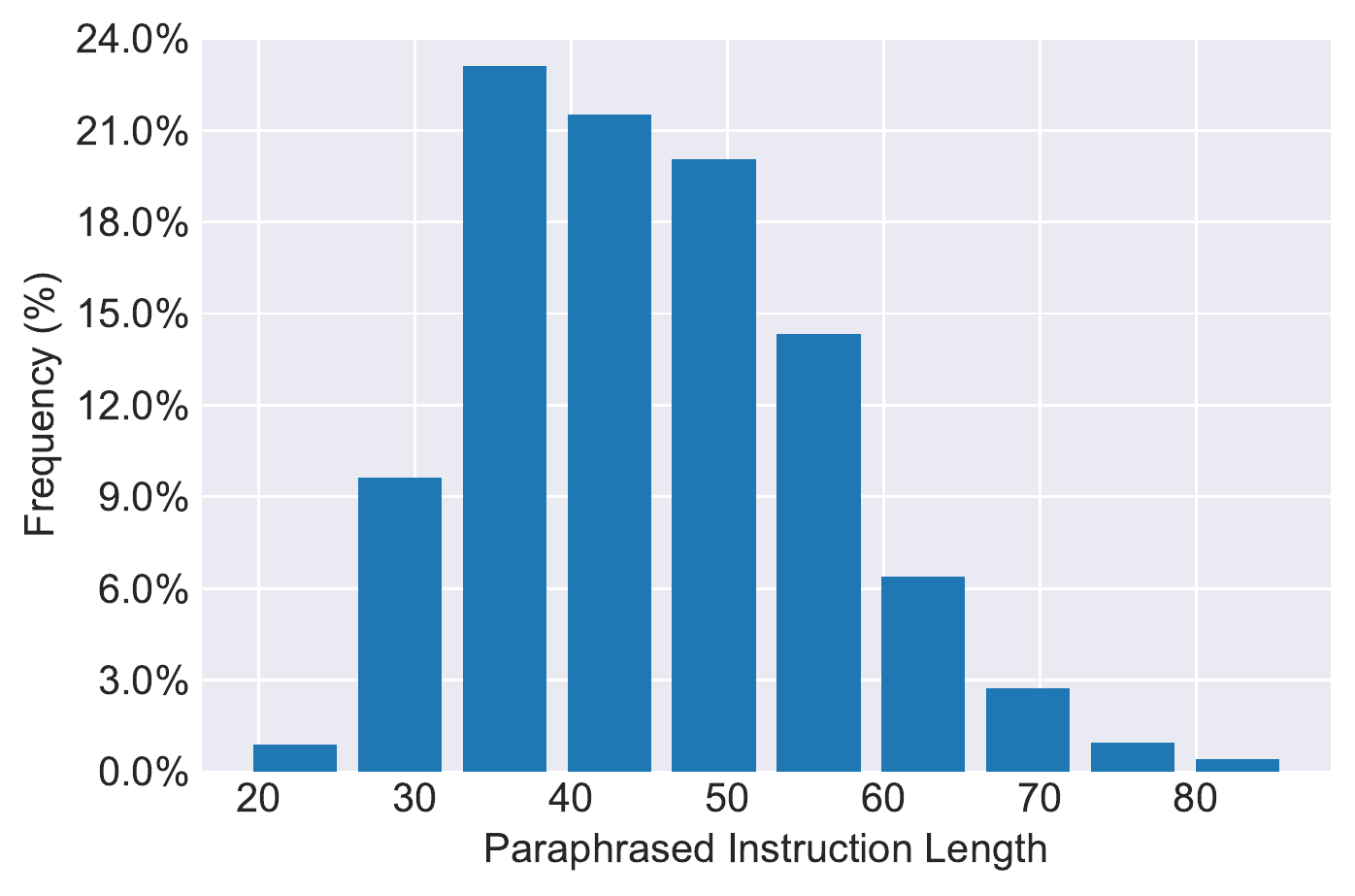}
    \includegraphics[width=0.325\textwidth]{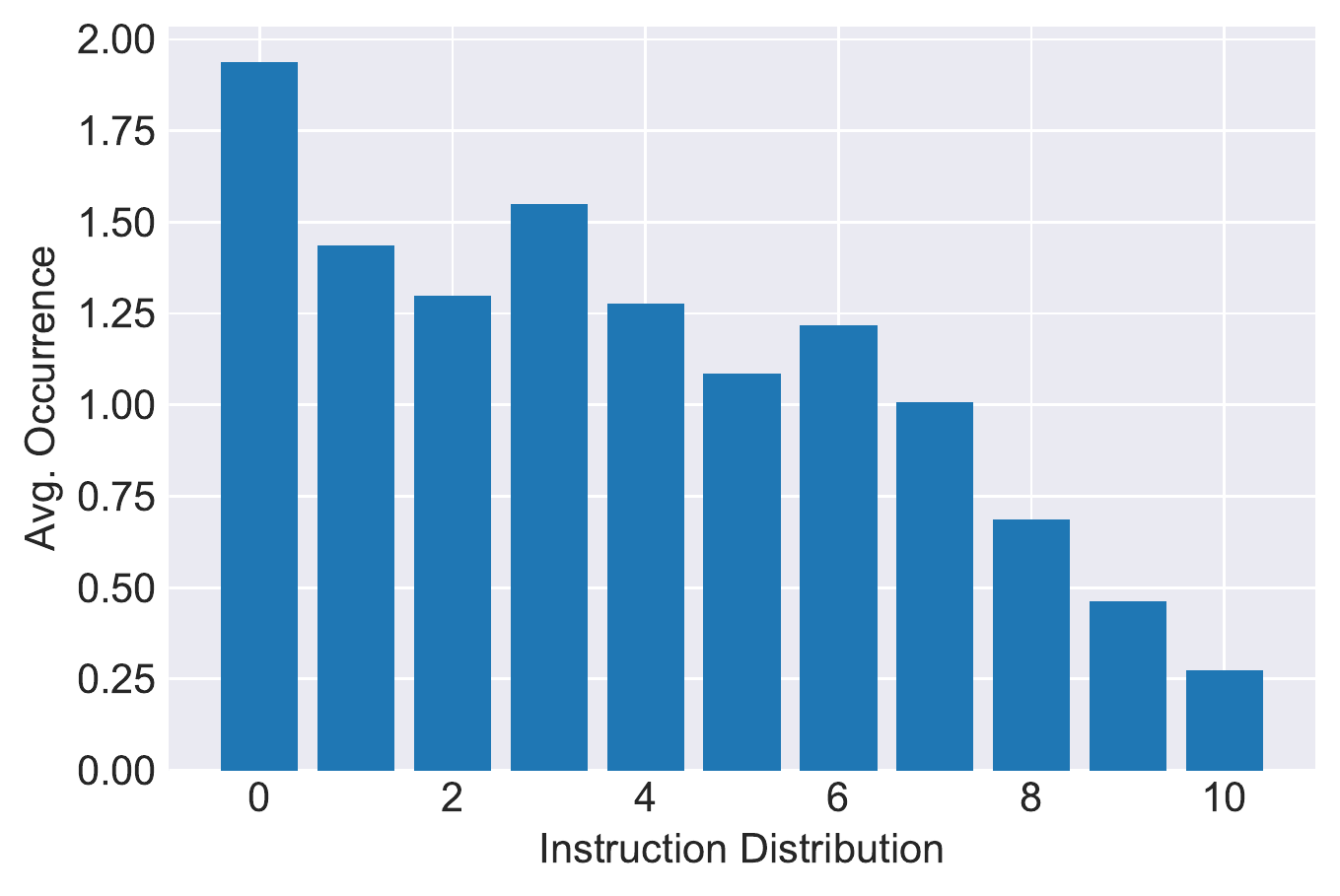}
    \includegraphics[width=0.325\textwidth]{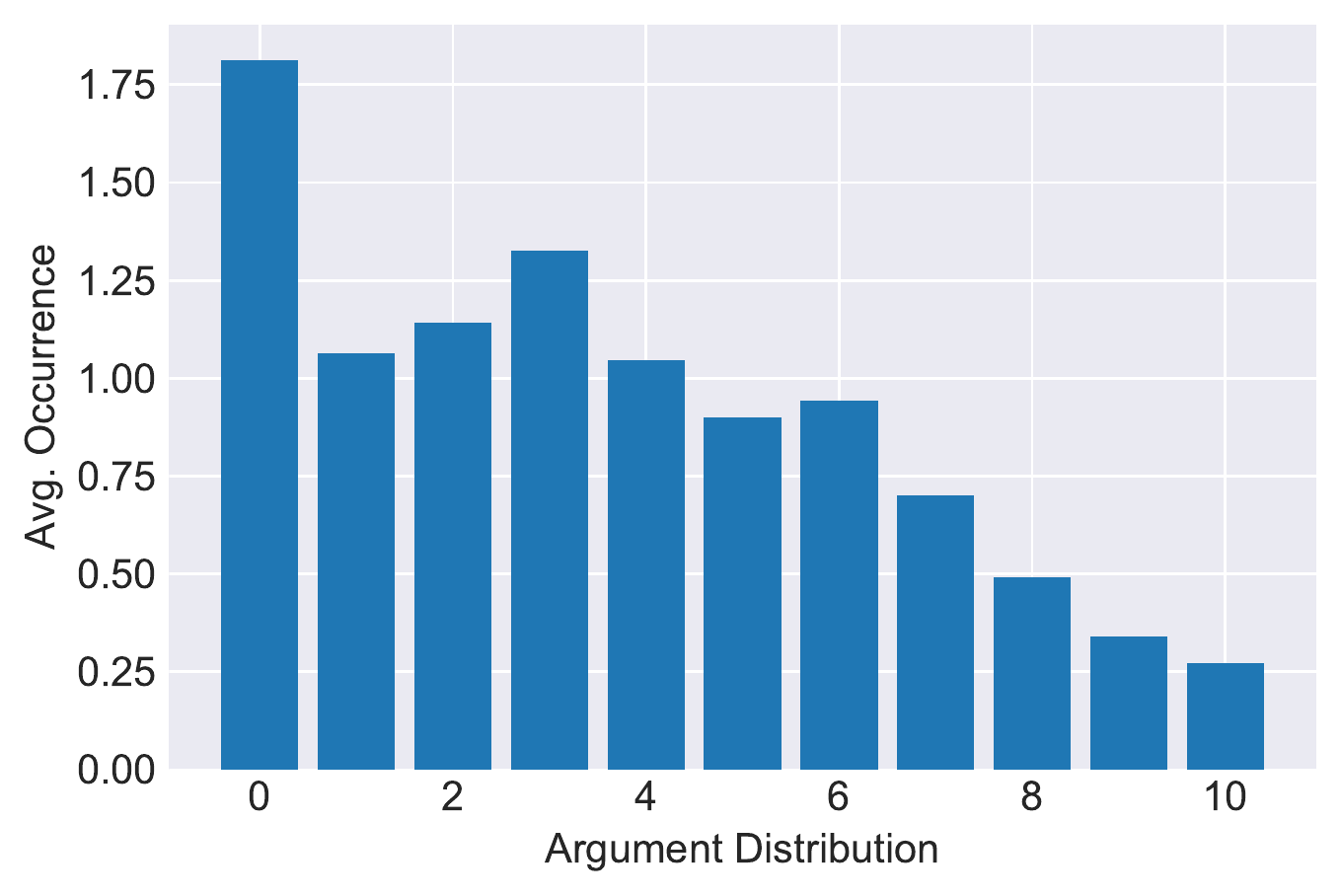}
    \caption{Distributions on dialogues and manuals of MagDial. User/Agent sentence length distribution (top left/top middle), instruction length distribution before/after paraphrasing (top right/bottom left), and distribution of number of instructions/arguments per turn (bottom middle/bottom right).}
    \label{fig:manual_stat}
\end{figure*}

Table \ref{tab:attr_type} presents the full list of attributes used in the database and manuals. Fig. \ref{fig:manual_stat} illustrates some data distributions on dialogues and manuals of MagDial. The average number of sentence length is 22.27 and 32.41 for the user and agent side, respectively. After the manual paraphrasing process, the average instruction length rises from 34.59 to 45.23, and its distribution becomes more normal. The number of instructions per turn shares a similar distribution with the number of arguments per turn since most instructions trigger one API call. Noticeably, neither an instruction is selected, nor an argument is triggered at a certain rate (around 20\%) of all dialogue turns since some turns only involve simple and generic dialogue behaviors such as greeting and farewells.

\subsection{Examples}

\begin{figure*}[!t]
    \centering
    \includegraphics[width=\linewidth]{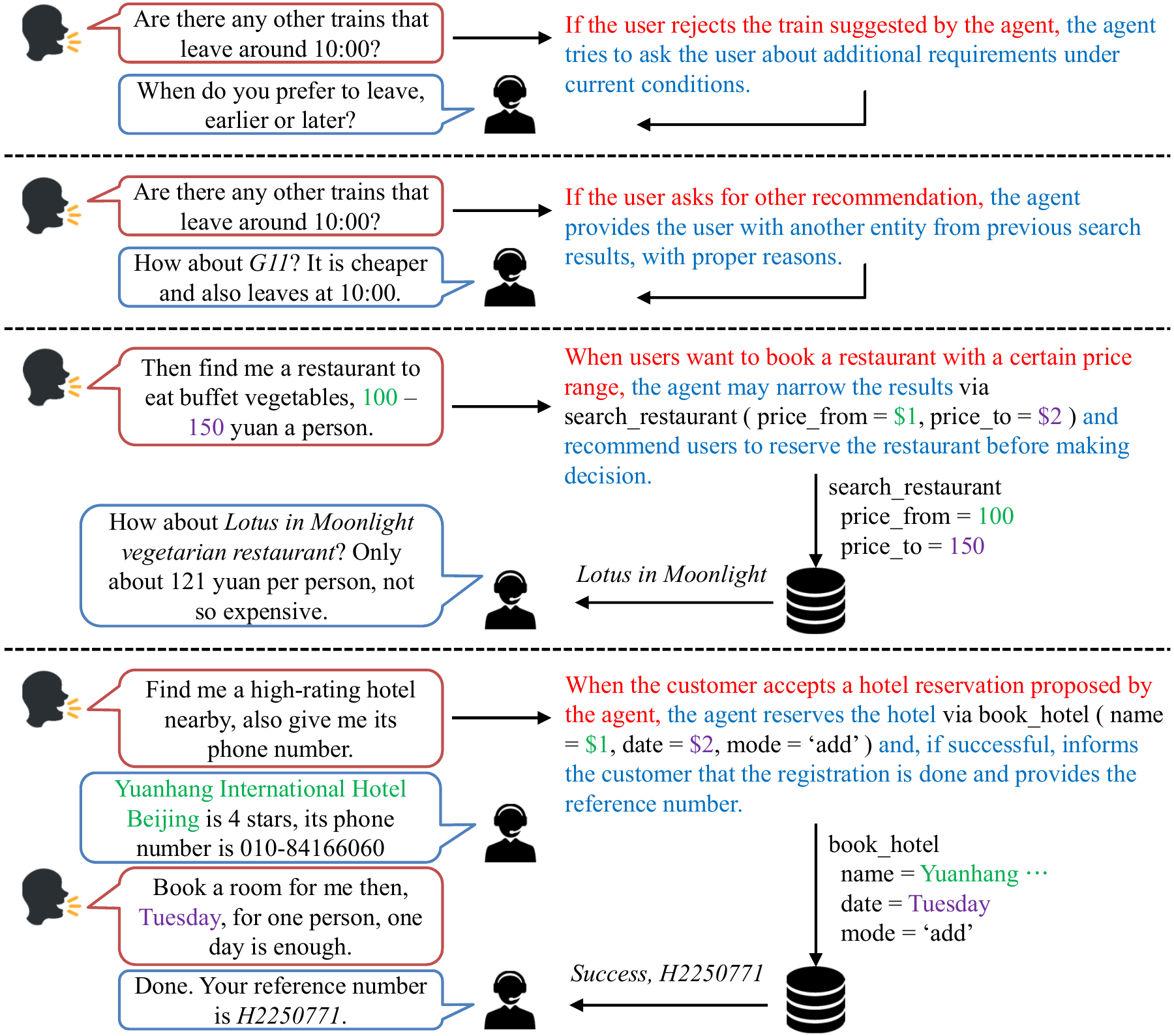}
    \caption{Several sample segments of conversation (left) in MagDial with selected instructions (right) in a manual. The paragraph marked in red/blue background color indicates the condition/solution part of an instruction. Each instruction may trigger an API call, and every API has its own input and output. As for argument filling, the agent detects the argument (\textit{e.g.}, Tuesday) from the dialogue context and recognizes its attribute type by reading the attribute mentions (\textit{e.g.}, date) in the selected instruction. The text spans to be extracted as an argument are marked in the same color with the corresponding attribute of APIs.}
    \label{fig:sample_all}
\end{figure*}

Fig.~\ref{fig:sample_all} presents four dialogue segments from the dataset. From sample segments, we can observe that the selected instructions summarize the current dialogue situation (the condition part) and provide the agent with some guidance on the next action (the solution part). The agent may choose different $I_t$ from $M$ at a turn $t$ because multiple responses can be appropriate for the same $D_t$, leading to different valid dialogue policies towards task completion. For example, in the first and second segments, the agent can either ask users what they require, or recommend another entity given the same user query, thereby producing different responses. The essence of the manual-guided dialogue scheme is to help the agent \textbf{understand} what the task is and how to act by reading manuals, so that the agent can better handle diverse user behaviors and agent policies instead of simply imitating a function mapping from conversation to response. Therefore, the agent guided by manuals can learn to interact with users \textbf{more efficiently}.

Most parts of the manual also include API calls (\textit{e.g.}, the third/fourth segment). The agent should follow the manual to extract arguments from dialogue context to interact with the database properly. At each turn $t$, the agent selects relevant $I_t$ by scanning each instruction on $M$, and extracts $A_t$ to be passed in API calls by detecting text spans from $D_t$. When predicting text spans, the agent reads the $I_t$ to focus on the target arguments, and does not need to recognize its attribute type beforehand. This indicates that all the annotations used in our scheme are domain-agnostic, thus improve the scalability across domains and tasks.

\subsection{Annotation Interface}

\begin{figure*}[!t]
    \centering
    \includegraphics[width=\linewidth]{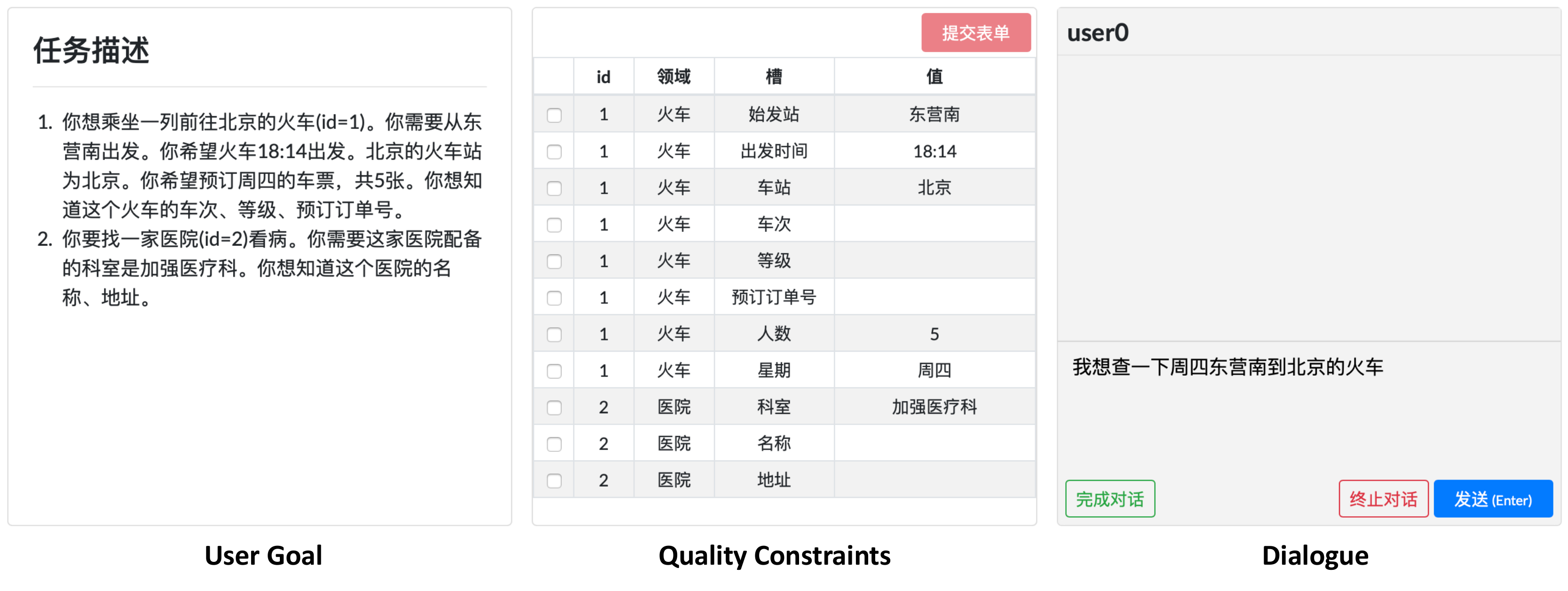}
    \caption{Interface from the user side, where the user goal description/quality constraints/dialogue is presented in the left/middle/right column respectively.}
    \label{fig:user_interface}
\end{figure*}

\begin{figure*}[!t]
    \centering
    \includegraphics[width=\linewidth]{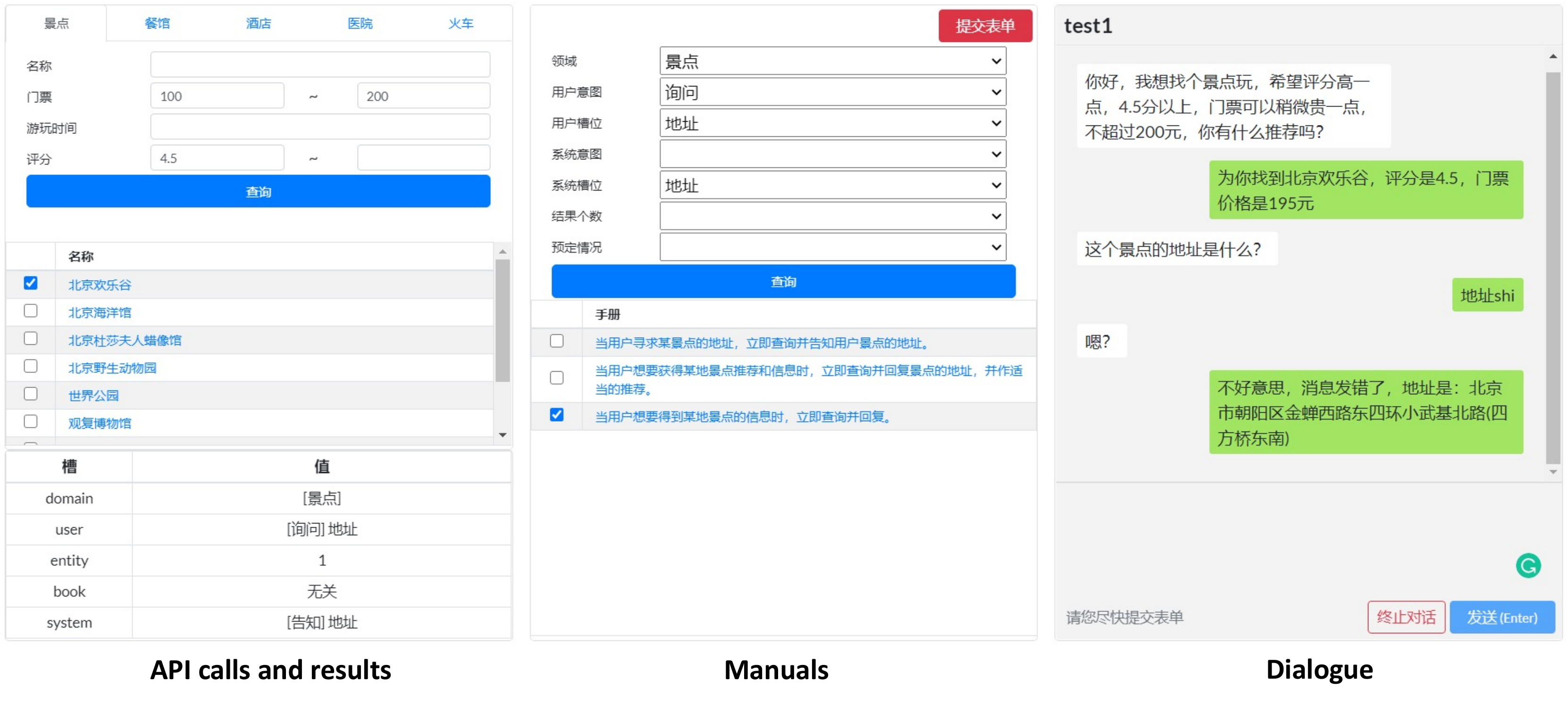}
    \caption{Interface from the agent side, where the API calls and results/manuals/dialogue is presented in the left/middle/right column respectively.}
    \label{fig:agent_interface}
\end{figure*}

Fig. \ref{fig:user_interface} provides the user side interface with a goal description (the left column) and a table listing constraints for quality control (the middle column). Fig. \ref{fig:agent_interface} presents the agent side interface with API calls and results (the left column) and manuals (the middle column).

\end{document}